\documentclass[journal,transmag, twocolumn]{IEEEtran}  % Comment this line out if you need a4paper

\IEEEoverridecommandlockouts                              % This command is only needed if 
\usepackage{graphics} % for pdf, bitmapped graphics files
\usepackage{epsfig} % for poillustratorstscript graphics files
\usepackage{amsmath} % assumes amsmath package installed
\usepackage{amssymb}  % assumes amsmath package installed
\usepackage{bm}  
\usepackage{subfloat}
\usepackage{algorithm,algorithmic}

\usepackage{graphicx}

\title{\LARGE \bf
	Nonparametric Inverse Dynamic Models for Multimodal Interactive Robots
}
\author{\IEEEauthorblockN{Kevin Haninger and Masayoshi Tomizuka}
	\IEEEauthorblockA{Department of Mechanical Engineering,
		University of California, Berkeley, CA 94720 USA}
}

\begin{document}

\IEEEtitleabstractindextext{%
\begin{abstract}
Direct design of a robot's rendered dynamics, as in impedance control, is now a well-established control mode for uncertain environments.  When the true interaction port variables are not measured directly (typically the generalized forces), the accuracy in reshaping the apparent dynamics is limited by that of the dynamic model which relates these interaction port variables to measured ones.  A typical example is serial manipulators with joint torque sensors, where the interaction occurs at the end-effector. This paper first examines the use of inverse dynamics in interactive control, examining causality and invertibility of traditional and general load dynamics. To perform increasingly complex tasks, these robots will be intermittently coupled with additional dynamic elements such as tools, grippers, or workpieces, some of which should be compensated and brought to the robot side of the interaction port. Furthermore, there may also be unavoidable and unmeasured external input when the desired system cannot be totally isolated. Towards semi-autonomous robots capable of handling such applications, a multimodal Gaussian process regression is developed.  As this regression approach returns not only an expectation, but a full distribution at each evaluation, a probabilistic means of identifying and separating externally perturbed data is proposed.  The passivity of the overall approach is shown analytically, and experiments examine the performance and safety of this approach on a test actuator. 
\end{abstract}}

\maketitle 

\section{Introduction}
\IEEEPARstart{I}{nteractive} robots in unstructured or collaborative environments achieve safety and productivity through reshaping the apparent dynamics of the robot. This manifests as taking external motion or force as input, and controlling the complementary variable according to desired dynamics, as popularized in impedance control \cite{hogan1985}.

\begin{figure}[h]
	\begin{centering}
		\includegraphics[width=\columnwidth]{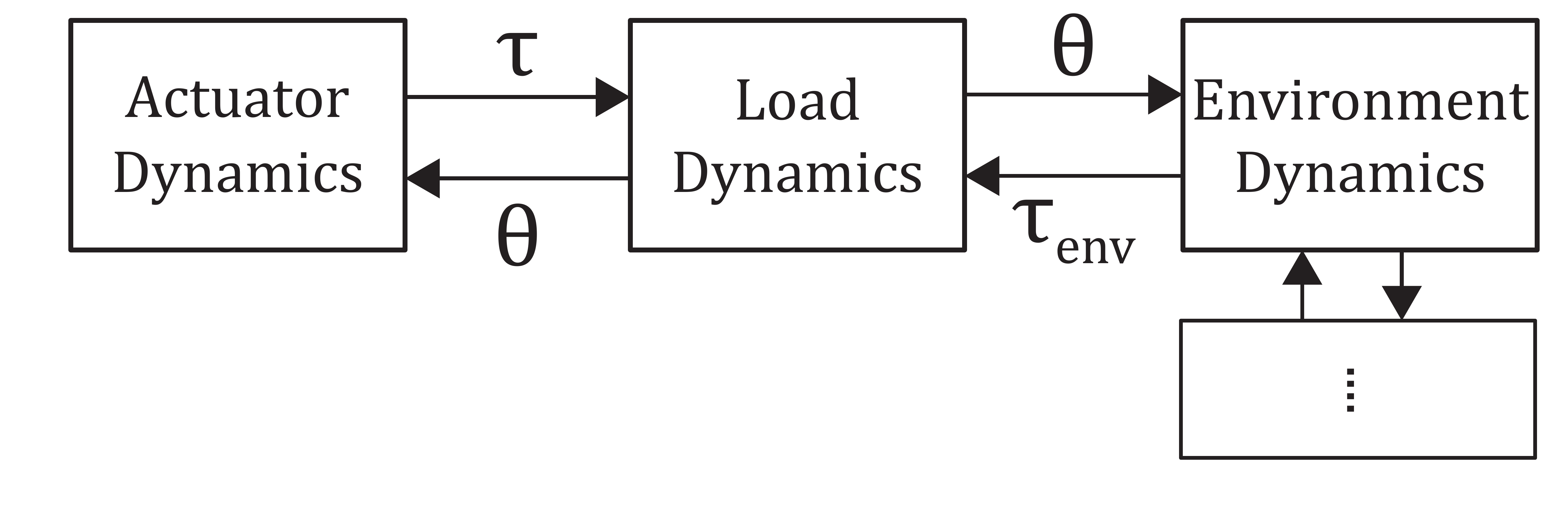} 
		\par
	\end{centering}
	\protect\caption{Interactive System Model \label{motor_load_env_model}}
\end{figure}

To realize the relationship between interaction port variables prescribed by the desired dynamics, these port variables must be either directly sensed or related to sensed variables through a dynamic model. In the latter approach, model accuracy limits the performance of the impedance control task.  For example, on many interactive robots joint-torque measurements are available (through torque sensors or series-elastic actuators), and it is often desired to realize interaction at the end-effector \cite{albu-schaffer2007, petit2015}. For these robots, the objective is to realize a control policy with sensed quantity $\theta$ and controlled response $\tau$, so as to render desired dynamics at  $\{\tau_{env},\theta\}$, as seen in Figure \ref{motor_load_env_model}. 

Coupling with the environment changes the effective load dynamics, containing potentially useful information. Environment dynamics encode information about the structure or intentions of the environment, which can be useful in determining appropriate robot behavior. For example, the use of a model for a human collaborator can be used to determine robot behavior, e.g.  \cite{peternel2016}, where the stiffness presented by a human induces a complementary stiffness on the robot. In other applications, it may be desired to bring these dynamics into the system model and compensate them so further input can be reacted to. For example, a robot compensating the weight of a workpiece allows a human collaborator to manipulate it more freely.

\begin{figure}[h]
	\begin{centering}
		\includegraphics[width=\columnwidth]{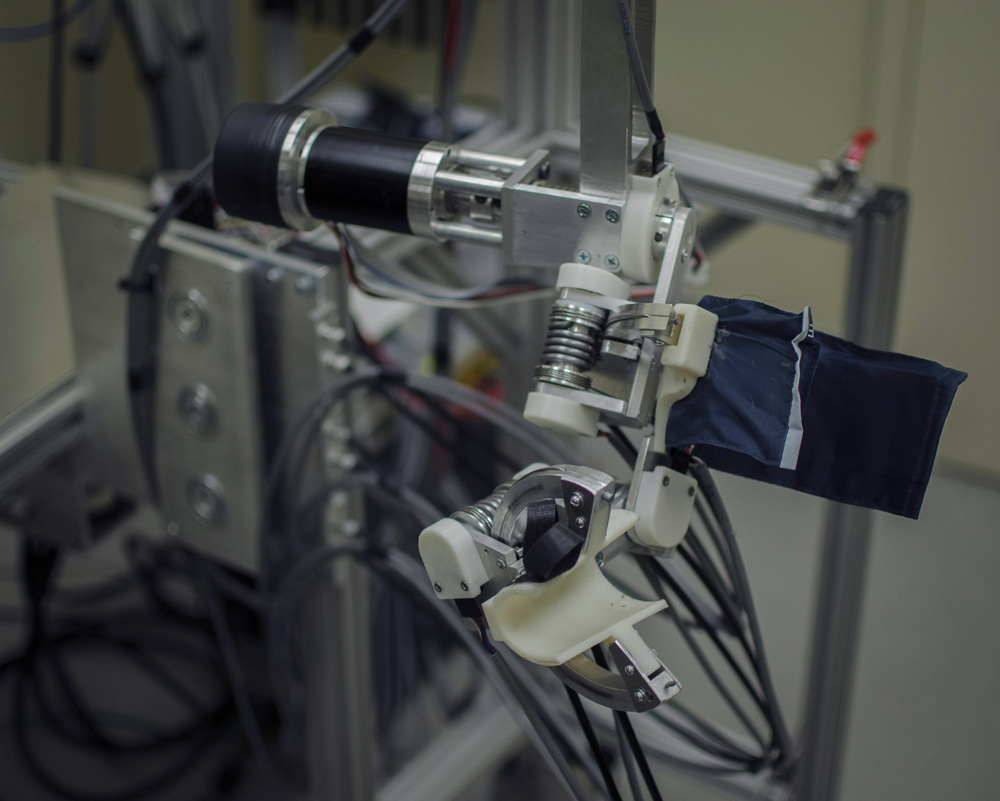} 
		\par
	\end{centering}
	\protect\caption{Upper-Limb Exoskeleton \label{exo_iso_view}}
\end{figure}

The learning of these effective load dynamics is useful (and also impacts safety \cite{haninger2016d}) but fundamental properties of interactive systems make this difficult. Many of these interactive robots present unconventional dynamics which limit traditional modeling approaches. For example, the exoskeleton shown in Figure \ref{exo_iso_view} has Bowden cables which exert an elastic force which resists their bending, and the subject wearing the exoskeleton introduces additional dynamics. 

Another modeling challenge arises from the difficulty in system isolation for interactive robots. All modeling techniques (which here includes model-free, data-driven techniques) rely on measurement of all inputs and outputs related by the system of interest. Currently, this is achieved by physical separation, such that all input to the system of interest arises from a controlled (or at least measured) source. However, as robots move from laboratories towards well-connected, real-world environments, isolation may be a limitation. In some applications, isolation is not possible for experimental reasons.  A subject wearing an exoskeleton may exert additional muscular torques which are not intrinsic characteristics of their body, but arise from complex, non-repetitive externalities. In other cases, it may simply be more convenient to allow additional input - such as manual excitement to induce free motion through specific regions of the state space during model identification.  

To varying degrees, all model identification rely on a roboticist applying a priori knowledge on model structure through design of experiment and identification technique to isolate and describe relevant parts of the system. Though mostly effective, this can be time consuming and fundamentally relies on an operator's a priori knowledge of the dynamics. To move towards autonomous, algorithmic treatment of multimodal dynamics, a unified framework is here developed to simultaneously classify a single data set into discrete dynamic modes, and model them. There is evidence that multimodal identification and switching also informs human motor control \cite{shadmehr1994}, suggesting this approach may eventually allow robots to achieve more general, human-like interaction outside the laboratory. 

In this paper, first the objectives of model learning in an interactive setting are stated, and conditions for meeting these objectives with general state-based modeling discussed.  Mixed Gaussian processes are used to develop an inverse dynamic learning approach which can separate intermittent disturbances as well as cluster and model multiple operational modes. The Gaussian process framework allows a natural expression of the model certainty, which informs the classification of the data.  The safety of this learning and resulting compensation are investigated, and it's passivity shown.  Experiments validate the performance and safety of the resulting implementation.

\section{Related Work}
Model learning has a rich history in the controls and robotics communities, and several techniques are now well-established for traditional robots: adaptive control, iterative learning control, and nonparametric learning.  As interactive robots must maintain performance even when perturbed from the nominal trajectory, here only state-space modeling techniques are considered.

Adaptive control uses input and output data of the system to (in some variants) realize on-line identification of model parameters \cite{narendra1995}.  However, adaptive techniques require a structured or parameterized model, making nonlinear friction and non-conventional dynamics difficult to compensate.

Nonparametric modeling techniques have found application to inverse dynamics learning \cite{nguyen-tuong2011}.  These models can generalize from historical data to new trajectories, although the extent to which this is practically achieved is both application and parameter dependent. Being constructed from historical data, they can capture more difficult non-linear effects and don't require a priori knowledge of model structure.

Multimodal identification has also been treated in some of these frameworks. Adaptive control has treated multimodal models \cite{narendra1995, narendra1997}, but again requires a priori knowledge on model structure for each mode.  Nonparametric models can be extended to be multimodal \cite{shi2005, haruno2001,meeds2006,rasmussen2002} or even continuously varying \cite{petkos2007}. 

Several authors have investigated inverse dynamic modeling for changing payload mass \cite{jamone2014, shareef2016}, one of the most common ways interactive robots become multimodal, but here considerations of modeling and interactive performance are taken (modeling limitations, passivity, external input), and validation is in interactive performance, not trajectory tracking.  There is also some prior work on classification of external perturbations, such as under cyclic system motion \cite{berger2014} or in collision \cite{golz2015}.  Here, perturbation is not identified with an existing model, but co-occurs during identification, and is separated from the underlying model within the proposed framework.

\section{Model Identifiability}
Almost invetiably in interactive control it is desired to relocate the interaction port, such as in Figure \ref{motor_load_env_model}, where desired dynamics are to be rendered between the load and environment based off an actuator controller policy.  This section will discuss challenges and ways of viewing this operation, then conditions necessary to achieve certain reshaping.   

\subsection{Causality}
As seen in Figure \ref{motor_load_env_model}, load dynamics, having at least an inertial component, are most rigorously viewed as an admittance - taking a force as input, and realizing a position as output.  As pointed out in \cite{hogan1985}, environments are more naturally viewed as admittances, for example kinematic constraints and nonlinear springs, which must take force as input and produce a position as output.

From an input-output systems perspective, compensating an admittance (the load dynamics) with an impedance control policy (on the actuator) is, in general, non-causal.  If the load was a pure inertia, this would require both a model and differentiation of the sensor signal.  This is not realizeable in a classical controls sense, and this uphill struggle against causality is reflected in many implementation challenges.  

\subsection{Invertibility}
Input-output system representations are the classical foundation of impedance and interactive control, but this section will view this challenge as dynamics invertibility (instead of causality).  To more generally discuss the direct use of a learned dynamic model, suppose a system with load dynamics expressed as
\begin{eqnarray}
\dot{x} = f(x)+g(x)u \label{ident_eqn}.
\end{eqnarray} 

Let this system model be invertible if
\begin{eqnarray}
\exists u=\hat{u}(x,\mathcal{D},\tilde{u}) ~\mathrm{s.t.}~ \dot{x} = \tilde{u}. \label{dyn_comp_goal}
\end{eqnarray}
where the compensation term $\hat{u}$ is a function of the current evaluation of $x$.  If this approach is data-driven, it will also be a function of historical data  $\mathcal{D} = \left\{u_1,x_1,\dot{x}_1, \dots,u_T,x_T,\dot{x}_T\right\}$.

Although in general the convergence of a regression $\hat{u}(x,\mathcal{D})$ to achieve \eqref{dyn_comp_goal} will rely on the properties of the regression technique used, it is also necessary that the mapping $\left\{\dot{x},x\right\}\rightarrow u$ is unique. Suppose that multiple inputs $u_1\neq u_2$, realized at the same state $x_1 = x_2$,  achieve the same instantaneous motion $\dot{x}_1 = \dot{x}_2$.   Then, by direct consequence $f(x_1)\neq f(x_2)$ - i.e. $f$ is not a function, and therefore cannot be exactly canceled and \eqref{dyn_comp_goal} cannot be achieved. 

Several factors can cause the loss of uniqueness for the system dynamics. One is the existence of hidden states or significant model limitations.  If these hidden states are slowly varying, this system may be better suited for adaptive or online techniques which capture the dynamics at the current operating condition. No known literature explores the use of such identification in interactive control.  This is in part due to a second source of non-unique dynamics: unobserved external input.

\begin{figure}[h]
	\begin{centering}
		\includegraphics[width=.6\columnwidth]{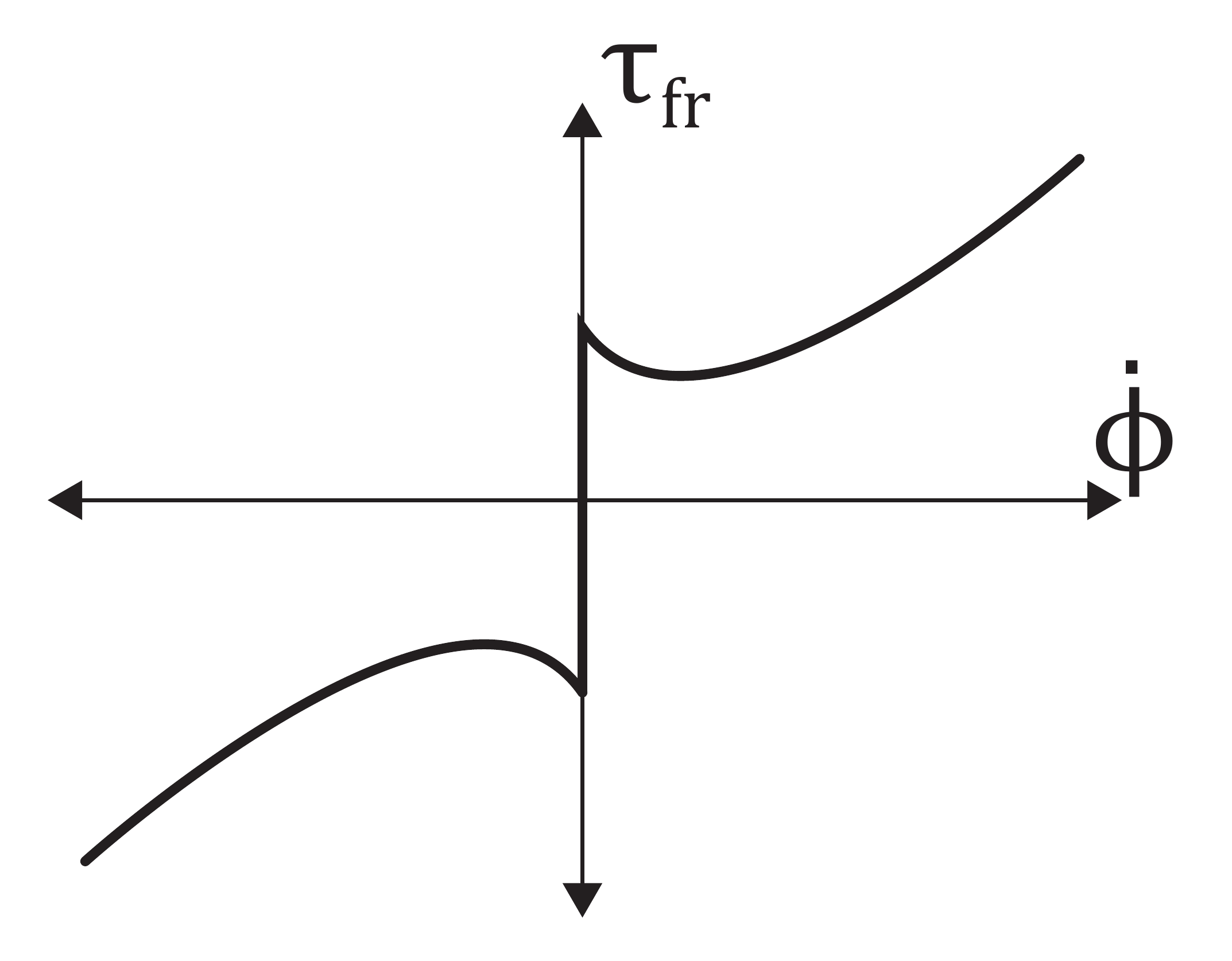} 
		\par
	\end{centering}
	\protect\caption{Friction Model with Coulomb Friction and Stribek Effect \label{friction_stribek}}
\end{figure}

Even if the system is isolated (no additional input) and well modeled (no hidden states), another significant source of non-uniqueness in real robots is coulomb or static friction, such as in Figure \ref{friction_stribek}. By being discontinuous at $\dot{\theta}=0$, such systems do not admit unique trajectories (i.e. the differential equations governing their behavior are not Lipschitz continuous). Any torque below the breakaway value will induce no motion. Techniques will be discussed later in the text to address this, but it remains a theoretical and practical limitation of truly model-free dynamics learning. 

\subsection{Load Dynamics}
Most implementations of impedance control implicitly reframe the problem for tractability. For theoretical treatments, issues of causality are typically avoided by assuming access to position, velocity and acceleration \cite{hogan1985}. In practice, often these higher order terms are dropped and limitations in the re-shaping of the dynamics accepted - effectively the impedance mass and damping will be those of the physical system.

A general schematic for an impedance controlled robot interacting with the environment is shown in Figure \ref{motor_load_env_model}. For now ignoring actuation dynamics, and taking $\tau$ as the controlled input to the system, the robot's inverse dynamics (load dynamics) can be written as
\begin{eqnarray}
M\left(\theta\right)\ddot{\theta} + C\left(\dot{\theta},\theta\right)\dot{\theta}+ g\left(\theta\right) & = & J_{int}^T\left(\theta\right)F_{int}+\tau. \label{rigid_robot_dyn}
\end{eqnarray}
The interaction force $F_{int}$, modulated by the Jacobian $J_{int}(\theta)$ is taken as the environmental input to the system. The objective is to realize desired Cartesian space or joint space dynamics of
\begin{eqnarray}
M_{imp}^j\ddot{\theta}+B_{imp}^j\dot{\theta}+K_{imp}^j\theta &=& J^T\left(\theta\right)F_{int} \label{joint_imp_ctrl} \\ 
J^{T}\left(\theta\right)\left(M_{imp}^c\ddot{x} + B_{imp}^c \dot{x}+K_{imp}^c x\right) &=& J^T\left(\theta\right)F_{int}. \label{cartesian_imp_ctrl}
\end{eqnarray}
where $[\cdot]_{imp}$ are the impedance parameters, and $x$ is the end-effector position achieved by the manipulator. These equations represent joint \eqref{joint_imp_ctrl} and Cartesian \eqref{cartesian_imp_ctrl} space impedance control  respectively. To achieve compliant trajectory tracking, $\theta$ or $x$ can be regarded as the deviation from the desired position.  

In early realizations of impedance control the entire robot dynamics were directly compensated (e.g. $\tau^d = -M(\theta)\ddot{\theta}...$) \cite{hogan1985}. Although in theory this allowed assignment of the apparent inertia, damping and stiffness of the manipulator, it required direct measurement of interactive force as well as acceleration, limiting realizability. Typical approaches to impedance control leave the desired impedance mass as the system's true inertia, and only the gravitational terms are compensated.  If this gravity compensation is done with a model $\hat{g}\left(\theta\right)$, the equilibrium equations of \eqref{cartesian_imp_ctrl} become:
\begin{eqnarray}
J^{T}\left(\theta\right)K_{imp} x\left(\theta\right)+\tilde{g}\left(\theta\right) &=& J^T\left(\theta\right)F_{int}. \label{cartesian_imp_err}
\end{eqnarray}
where $\tilde{g}$ is the error in the gravitational model $\tilde{g} = g-\hat{g}$. This error directly affects the equilibrium conditions just as an external force. As the stiffness $K_{imp}$ decreases, a large deviation from desired position can occur. This sets in-practice limitations to the impedances which can be rendered on a robot, and due to the position dependency of the model error, can give unexpected performance in different parts of the joint space.

\subsubsection{Joint Flexibility}
Many interactive robots present joint flexibility induced by the joint-torque measurement devices.  This seems inherent to torque measurement (strain/displacement is instrumentable, torque is not directly), with typically an inner-loop torque control being used to improve the backdriveability of the system.   Although this hierarchical control presents practical improvements for interactive performance, it raises questions about where, and how to inject a learned model.  Ostensibly, a model could be developed over the states of the motor and load, and directly added at the motor torque (i.e. direct injection).  This is not pursued here for several reasons.  The increase in state space from including the two additional states for the motor may be prohibitive.  Guarantees of safety through passivity become much harder to demonstrate.  Furthermore, the dynamics of the motor include significant coulomb friction, making the invertibility required for high-performance dynamic inversion tenuous.  Thus, here, well-established inner-loop control design methodologies, which can suggest the safety and performance a priori are used. 

\section{Generalized Load Dynamics\label{learning_dynamics}}
\subsection{Multimodal Dynamics Learning}
For a multimodal system, the dynamics are assumed to switch between multiple dynamic models; i.e. $\tau=f_{k}\left(\ddot{\theta}, \dot{\theta},\theta\right)$ where $k\in\left[1,\dots K\right]$ indexes the possible dynamic modes. Given a data collection $\mathcal{D}=\left\{ \tau_{1},x_{1},\dots,\tau_T,x_T\right\}$, where here $x_t=\left[\ddot{\theta}_t, \dot{\theta}_t,\theta_t\right]$, assume that these samples are independently and identically drawn as
\begin{eqnarray}
\tau_{t}\sim p\left(\tau|x_{t},w_{t},\Theta\right)
\end{eqnarray}
where $w_{t}\in\left[1,\dots,K\right]$ is a latent indicator variable showing the membership of sample $t$, and $\Theta=\left\{ \Theta_{1},\dots,\Theta_{K}\right\}$, where $\Theta_k$ parameterizes the $k^{th}$ distribution. 

Here, Gaussian processes are used to model each mode's inverse dynamics \cite{rasmussen2006}. Other works explore improvements for inverse dynamic modeling by using Locally Weighted Projection Regression \cite{vijayakumar2000}, Incremental Support Vector Machine \cite{ma2003} and Infinite Mixture of Linear Experts \cite{damas2013}, but the simplicity of Gaussian processes allows this paper to investigate phenomena which is largely invariant to regression technique. 

Under this regression technique, it is assumed that each data point takes the following distribution:
\begin{eqnarray}
p\left(\tau_t | x_t ,w_t = k, \Theta \right) & = & \mathrm{GP}\left(x_t,\mathcal{D}_k,\Theta_k \right) \\
& = & \mathcal{N}\left(\mu_t,\Sigma_t\right).
\end{eqnarray}
where $\mathrm{GP}$ denotes the posterior distribution of the Gaussian Process, based off data $\mathcal{D}_{k}=\left\{ \left\{ \tau_{t},x_{t}\right\} :\,\,w_{t}=k\right\}$, the subset of data labeled to this dynamic mode. This evaluates to a normal distribution, where $\mu_t$ and $\Sigma_t$ are implicitly functions of the mode data $\mathcal{D}_k$ and parameters $\Theta_k$. 

Under a slight abuse of notation (distribution will also be conditioned on the state) the clustering and parameter fitting can be framed as a maximum likelihood problem
\begin{eqnarray}
\max_{\Theta}\,\,p\left(\mathcal{D}|\Theta\right)=\sum_{w\in W} p\left(\mathcal{D}|\Theta,w\right)p\left(w\right) \label{max_lik_overall}
\end{eqnarray}
where $W$ is the set of all admissible combinations of clusterings. The expected complete likelihood and membership probabilities can be written in a straight forward application of the Expectation-Maximization algorithm:

\begin{eqnarray}
q\left(w|\mathcal{D},\Theta\right) &\propto& \prod_k p(\mathcal{D}_k |w, \Theta_k) p(w)\label{exp_dist}\\
\langle l\left(\Theta; \mathcal{D}, w\right)\rangle_q &=& \sum_{w\in\mathcal{W}} q(w|\mathcal{D}, \Theta)\log p\left(\mathcal{D},w|\Theta\right) \label{exp_log_lik}
\end{eqnarray}
where $\mathcal{D}_k$ is the partition of the data belonging to the $k^{th}$ mode.  Note that neither step of the EM algorithm is computationally feasible. The set of possible clusterings $W$ is of size $K^T$, which will be prohibitively large for even modest data sets. This is inherent to nonparametric models, where the likelihood of a sample's latent mode depends on the classification of other data, as the mode membership is not independent conditioned on the parameters $\Theta$. To address this issue, others (e.g. \cite{shi2005}) use Markov-Chain Monte Carlo techniques to sample from the distribution. 

Some MCMC variants of EM sample from the distribution $q(w|\mathcal{D},\Theta)$ to approximately marginalize the expected log likelihood \eqref{exp_log_lik}. The limiting case of pulling a single sample from $q^{(n)}$ has been formalized as the Stochastic Expectation Maximization (SEM) algorithm \cite{celeux1992}, where demonstrations of convergence (in probability) can be found therein. Under this approach, let $w^{(n)}$ denote a sample from the current parameter estimates $\Theta^{(n)}$, and the sample likelihood written as:
\begin{eqnarray}
l_{w^{(n)}}(\mathcal{D}|\Theta) = \sum_k \sum_{t\in\mathcal{T}_k} \log\left(p\left(y_t |\mathcal{D}_k, \Theta_k, w^{(n)}_t\right) \right)p(w^{(n)}|\mathcal{D},\Theta^{(n)})
\end{eqnarray}
where $\mathcal{T}_k$ is the set of time indices which are labeled to the $k^{th}$ mode.

\subsection{Sampling Latent Class Membership $w$}
Direct sampling from \eqref{exp_dist} is again not computationally feasible. Generating these samples can be done with Gibb's sampling, alternatively viewed as a leave-one-out clustering approach. This gives conditional distributions of
\begin{eqnarray}
p(w_t | \mathcal{D}, w_{-t},\Theta) & \propto & p(w_t, y_t | \mathcal{D}, w_{-t}, \Theta) \\
& = & p(y_t | \mathcal{D}, w, \Theta) p(w_t | w_{-t}) \label{Gibbler}
\end{eqnarray}

The evaluation of $p(y_t | \mathcal{D}, w, \Theta) = \mathrm{GP}(x_t,\mathcal{D}_k,\Theta_k)$ is a straightforward evaluation of the distribution returned by the Gaussian process.  If the mode $w_t$ is assumed to be i.i.d., the conditional distribution $p(w_t|w_{-t})=p(w_t)$ is trivial to evaluate. However, different priors for $w_t$ can be applied to characterize the mode switching behavior. If additional sensing gives inference to the mode, this can be incorporated by making this distribution conditional on the new data $\mathcal{D}'$, as $p(w|\mathcal{D}')$.  Here, a simple time correlation will be assumed as:
\begin{eqnarray}
p\left(w_{t}|w_{t-1}\right)=
\begin{cases}
\pi & w_{t}=w_{t-1}\\
1-\pi & w_{t}\neq w_{t-1}
\end{cases}
\end{eqnarray}

This gives overall probability
\begin{eqnarray}
p\left(w\right)	& = & \pi^{c_{0}}\left(1-\pi\right)^{T-c_{0}} \\ 
c_{0} & = & \sum\mathbb{I}\left(w_{t}=w_{t-1}\right).
\end{eqnarray}
Where $\mathbb{I}$ is an indicator function of one when it's argument is true and zero otherwise. This can be easily normalized to find $p(w_t|w_{-t})$ as needed for \eqref{Gibbler}.

\subsection{Parameter Identification}
Straightforward maximum likelihood estimation of (hyper-)parameters $\Theta$ is again not tractable due to the coupling of the latent variables. Given a sample of the classification $w^{(n)}$, parameters can be estimated as follows:
\begin{eqnarray}
\langle l(\mathcal{D}|\Theta) \rangle_{w^{(n)}}= \sum_{k} l\left(\mathcal{D}_k|\Theta_k\right)p(w^{(n)}|\mathcal{D},\Theta^{(n)}) \label{exp_lik_SEM}
\end{eqnarray}
where the data partions $\mathcal{D}_k$ are induced by $w^{(n)}$.  Although the total marginal likelihood of a set of data drawn from a Gaussian process can be easily computed (e.g. \cite{rasmussen2006}), here the hold-one-out likelihood is used; as in practice this has given better performance:
\begin{eqnarray}
l\left(\mathcal{D}_k|\Theta_k\right) = \sum_{t\in\mathcal{T}_k} -\left(\tau_t-\mu_t\right)^T \Sigma_t^{-1}  \left(\tau_t-\mu_t\right)^T
\end{eqnarray}
where $\mu_t$ and $\Sigma_t$ the posterior mean and covariance returned by evaluating gaussian process $\mathrm{GP} \left(\mathcal{D}_k, \Theta^{(n)}_k\right)$. As the parameters $\Theta_k$ are all scalar, they can be searched over with a numerical gradient descent in each iteration of the algorithm.

\section{Disturbance Identification \label{ext_dist}}
This framework can also be used to rigorously motivate a means of separating intermittent disturbance from model uncertainty.  Let the general dynamic system from \eqref{ident_eqn} be extended to include unknown input $d$.  The system dynamics become:
\begin{eqnarray}
\dot{x} = f(x)+g(x)\left(u+d\right)\label{ident_sys} 
\end{eqnarray}

The exogenous input $d$ is not measured, but by being additive with $u$ will affect system evolution. As this input is unmeasured (but it is presumed that $u$, $x$, and $\dot{x}$ are), this will cause an instantaneous deviation in $\dot{x}$, leading to an evaluation of the regression function equivalent to $\left\{\dot{x}_d,x \right\}\rightarrow u$, where $\dot{x}_d = \dot{x}-g(x)d$. If the regression function is unique in the first argument, i.e. $\dot{x}_1\neq\dot{x}_2 \Rightarrow\hat{\tau}\left(\dot{x}_1,x\right)\neq\hat{\tau}\left(\dot{x}_2,x\right)$, this perturbation will cause an evaluation which differs from the measured input.   Typical regression techniques assume some distribution on samples $u_t$ to accommodate noise and regression error, however if $u_d$ significantly diverges from the $u$, it is likely the result of an external perturbation. One advantage of Gaussian Processes is the full posterior distribution of a sample is returned.  Practically speaking, if a point in the state-space has given consistent data, the posterior covariance will be smaller and perturbations more easily identified.

Note that if there is a state dependency of $d$, i.e. $d=\tilde{f}(x)$, ideologically, it could be more accurately viewed as model uncertainty. Such model uncertainty is then captured with the regression, and $u_d\rightarrow u$ as this disturbance's effect is learned by the regression technique. This motivates the practical and ideological condition that $d_k\perp x_k$, the disturbance should be independent of the state.  Conversely, the states define the system.  This can be easily obfuscated in interactive control, where additional systems are coupling with the robot. However, for many tasks, including the typical goal of compensating a tool's weight, the measured position can appropriately partition robot from environment.

Under assumptions of the exogenous input's time-series characteristics, perturbed data can be removed under the multimodal framework introduced above. To formalize the distinguishing of nominal and perturbed behavior, let the dynamics be as shown in \eqref{pert_dyn}.
\begin{eqnarray}
\tau_{t} & = & \begin{cases}
f\left(x_{t}\right) & w_{t}=1\\
f\left(x_{t}\right)+d_{t} & w_{t}=2
\end{cases} \label{pert_dyn}
\end{eqnarray}
where $f\left(x_t\right)$ is the inverse dynamics and disturbance $d_t$. If $f$ is fit with a gaussian process, each evaluation of $f(x_t)$ will yield a conditional distribution of $\tau_{t}$. Let $p\left(\tau_{t}|x_{t},w_t=1,\Theta\right)\sim\mathcal{\mathcal{N}}\left(\mu_{t},\Sigma_{t}\right)$, where $\mu_{t}$ and $\Sigma_{t}$ are the mean and covariance returned by the Gaussian Process. 

If the disturbance is assumed to be i.i.d. with $d_t\sim\mathcal{N}\left(0,\Sigma_{d}\right)$. This gives the distribution 
\begin{eqnarray}
p\left(\tau_{t}|x_{t},w_t=2,\Theta\right)\sim\mathcal{\mathcal{N}}\left(\mu_{t},\Sigma_{t}+\Sigma_d\right)
\end{eqnarray}
When the system is perturbed, it will deviate from the nominal dynamics, here characterized by the additional covariance in it's distribution.

\subsection{Disturbance Parameter Identification}
An appropriate value of $\Sigma_d$ is key to the performance of sampling and parameter updating as in \eqref{Gibbler}. Building off the SEM framework used in \eqref{exp_lik_SEM}, let the identification be done by maximizing the following expected likelihood: 
\begin{eqnarray}
\langle l\left(\mathcal{D}_2|\Sigma_d\right) \rangle_{w^{(n)}}=  \sum_{t\in\mathcal{T}_2}  -p(w^{(n)}_t | \mathcal{D}, \Theta^{(n)})~\tilde{\tau_t}^T\left(\Sigma_{t} + \Sigma_d\right)^{-1}\tilde{\tau}_t
\end{eqnarray}
where $\tilde{\tau}_t=\tau_t - \mu_t$, $\mu_t$ and $\Sigma_t$ are the posterior mean and covariance of gaussian process $\mathrm{GP}\left(\mathcal{D}_1,\Theta^{(n)}_1\right)$ evaluated at $x_t$. Again, if $\Sigma_d$ is low-dimensional (e.g. $\Sigma_d \propto I$), this can be easily searched over numerically.

\section{Passivity in Feedforward Dynamics Compensation}
For interactive robots, the safety must be guaranteed in interactive tasks, not just stability in isolation (e.g. Lyapunov approaches). The most common approach in interactive control is to show the passivity of the robot \cite{hogan1985, albu-schaffer2007}. If the robot and controller are passive, they can be coupled to an arbitrary environment - payload, working surface, without compromising coupled system stability.  The intuitive interpretation of a passive system is one which has a lower bound on the energy which can be extracted through the interaction port. However, feedforward techniques can even problematic for stability, much less passivity. 

In particular, friction compensation can be dangerous. Viscous friction compensation, e.g  a term of the form $\tau = \hat{\beta}\dot{\theta}$ is directly injecting power into the system ($\tau\cdot\dot{\theta}>0$), and if $\hat{\beta}$ exceeds the real viscous friction, system energy will grow without bound.  Furthermore, physical damping is key to providing robustness in interactive control when discretization or model uncertainty is considered \cite{colgate1994, haninger2016d}. 

These stability considerations are why the introduction of feedforward control in, e.g. iterative learning control, is done such that guarantees of system behavior under the feedforward control policy can be made.  This is a major challenge to model-free or nonparametric learning approaches, which by not having a model, limit the a priori analyses which can be undertaken. Here, let the robot dynamics be as shown in \eqref{rigid_robot_dyn}, with a control policy of:

\begin{eqnarray}
\tau = \tau_{ff}\left(\theta\right) + \tau_{imp}\left(\theta, \dot{\theta}\right)
\end{eqnarray}

A system with state $x$, dynamics $\dot{x}=f(x,u)$  and output $y=g(x)$, can be shown to be passive with respect to $\left\{y,u\right\}$ if $\exists$ a storage function $S(x)~:~\mathcal{X}\rightarrow\mathbb{R}$ which is bounded from below, and $S_x(x)\cdot f(x,u)<0$, where $S_x$ denotes the partial derivative of $S(x)$ with respect to $x$.

When $\tau_{ff}(\theta)$ is chosen from a Gaussian process using a square exponential kernel,
\begin{eqnarray}
k(x_i,x_j)= \sigma_y\exp\left(-\Vert x_1-x_2\Vert^2_2 l^{-2}\right)+\sigma_n\delta_{ij}
\end{eqnarray}
where $\delta_{ij}=1 \iff i=j$ and zero mean function, the feedforward torque can be written as:
\begin{eqnarray}
\tau_{ff}(\theta_t) &=& K_t^T K_{\mathcal{D}}^{-1}{\bm {y}} \\
K_t &=& \left[\begin{array}{c}
\sigma_{y}\mathrm{exp}\left(\mathtt{-}\left(\theta_{t}\mathtt{-}\theta_{1}\right)^{T}\left(\theta_{t}\mathtt{-}\theta_{1}\right)l^{\mathtt{-}2}\right)\\
\vdots\\
\sigma_{y}\mathrm{exp}\left(\mathtt{-}\left(\theta_{t}\mathtt{-}\theta_{T}\right)^{T}\left(\theta_{t}\mathtt{-}\theta_{T}\right)l^{\mathtt{-}2}\right)
\end{array}\right]
\end{eqnarray}
where $\sigma_y$ and $l$ are parameters of the GPR, and $K_\mathcal{D}$ denotes the kernel induced by data set $\mathcal{D}$, and $\bm {y}$ is the previously observed values in $\mathcal{D}$. 

Let the following function be defined:
\begin{eqnarray}
\mathcal{E}\left(x\right)&=&\left[\begin{array}{c}
\sqrt{2\pi}\sigma_{y}^{-1}l^{2}\mathrm{erf}\left(-\frac{\left(x-x_{1}\right)^{T}\left(x-x_{1}\right)}{l^{2}\sqrt{2}}\right)\\
\vdots\\
\sqrt{2\pi}\sigma_{y}^{-1}l^{2}\mathrm{erf}\left(-\frac{\left(x-x_{T}\right)^{T}\left(x-x_{T}\right)}{l^{2}\sqrt{2}}\right)
\end{array}\right]
\end{eqnarray}
where $\mathrm{erf}$ is the Gauss error function. By construction $\frac{\partial\mathcal{E}(x)}{\partial x} = K_t$. As the error function is bounded (above and below), any finite-valued training data means that a storage function $S_f(\theta)= \mathcal{E}(\theta)K_\mathcal{D}{\bm {y}}$ is bounded from below.

Note that the passivity of two systems in parallel can be directly concluded from the passivity of each. As the impedance controller is by construction passive, only the passivity of the feedforward term will be shown below.  Let the following storage function be defined:

\begin{eqnarray}
S(\theta,\dot{\theta}) & = & \frac{1}{2}\dot{\theta}^T M\left(\theta\right) \dot{\theta} + \mathcal{E}(\theta){K}_\mathcal{D}{\bm y}+ V_g(\theta) \\
\dot{S}\left(\theta,\dot{\theta}\right) & = & -\dot{\theta}^T\left(\dot{M}\left(\theta\right)-2C(\theta,\dot{\theta})\right)\dot{\theta} + \dot{\theta}^TJ^T\left(\theta\right)F_{int}
\end{eqnarray}
where $V_g\left(\theta\right)$ is the gravitational potential energy of the system, such that $\frac{\partial V_g(\theta)}{\partial \theta}= g(\theta)$. As $\left(\dot{M}\left(\theta\right)-2C(\theta,\dot{\theta})\right)$ is skew-symmetric, this quadratic term is zero. Thus we achieve $\dot{S}\leq\dot{\theta}^T\tau_{int}$, and the robot is passive.

Note that there are some limitations to this passivity approach. It shows passivity for only regressions of the joint angles, although this is often sufficient for many interactive robot applications, and may even be desired to prevent over-compensation of friction. However, to regress only on position breaks the conditional independence of $\tau_t$. Fortunately, the overall regression technique can still include higher-order terms, it is just $K_t(\theta)$, the evaluation dependent on current state, which can only depend on $\theta$.  In the implementation here, this reduces to the simple evaluation of the regression at $\hat{\tau}(0,0,\theta)$.

A further limitation of this passivity argument is that it is only valid when an offline dataset $\mathcal{D}$ is used for inference, with no obvious extensions to a general online learning strategy.

\section{Experimental Results}
\begin{figure}[h]
	\begin{centering}
		\includegraphics[width=\columnwidth]{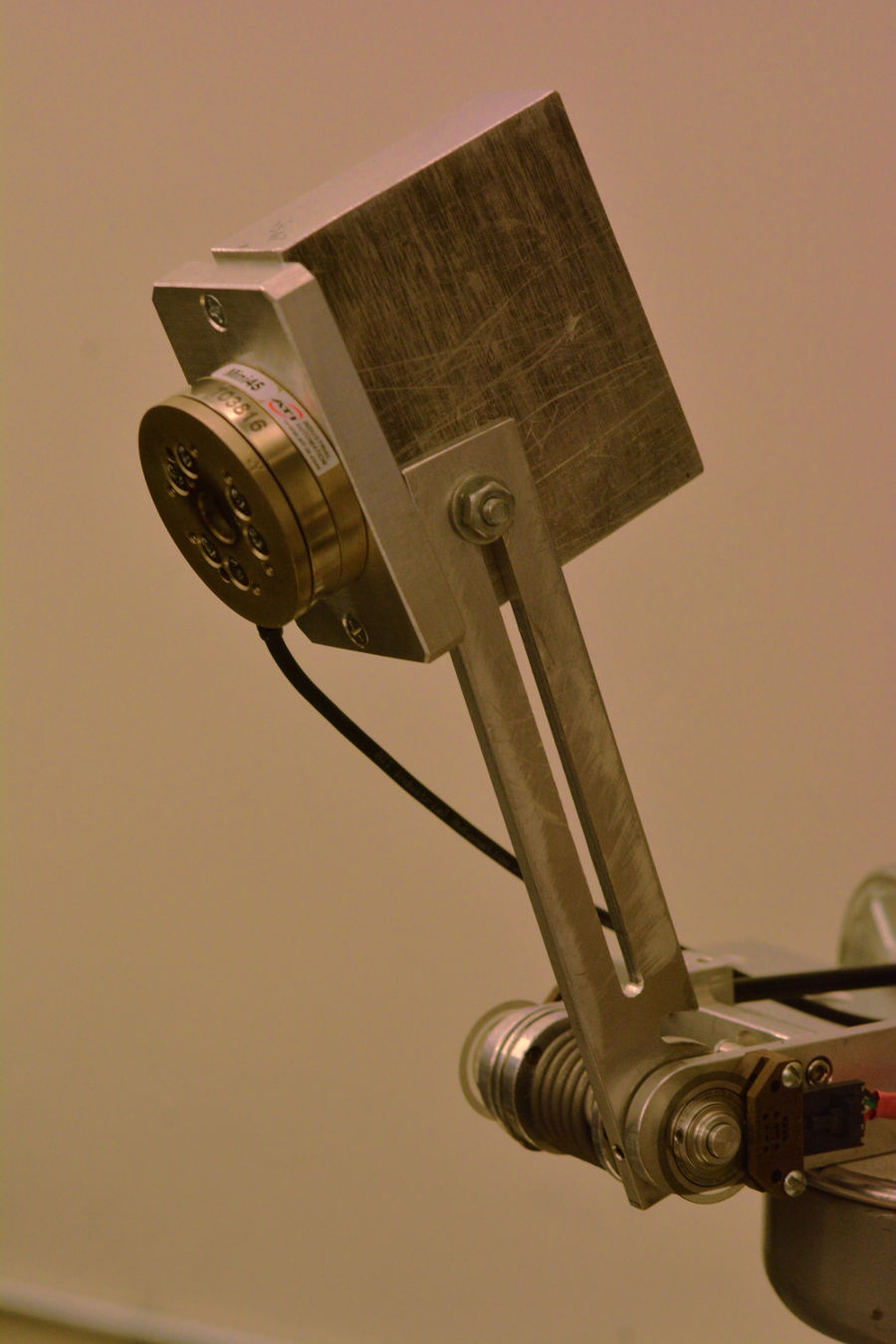}
		\par\end{centering}
	\caption{Experimental Setup \label{test_setup_vert}}
\end{figure}

The efficacy of this approach is first validated on a test actuator, seen in Figure \ref{test_setup_vert}.  An integrated force sensor allows validation of the mode classification, and the inertia of $.73 Kg$ acts as the load side dynamics.
Data collection is done with a high-gain PD position control in quasi-static exploration of the range-of-motion.  Data was downsampled to 20Hz for the model fitting. 

\subsection{Classification}
\begin{figure}[h]
	\begin{centering}
		\includegraphics[width=\columnwidth]{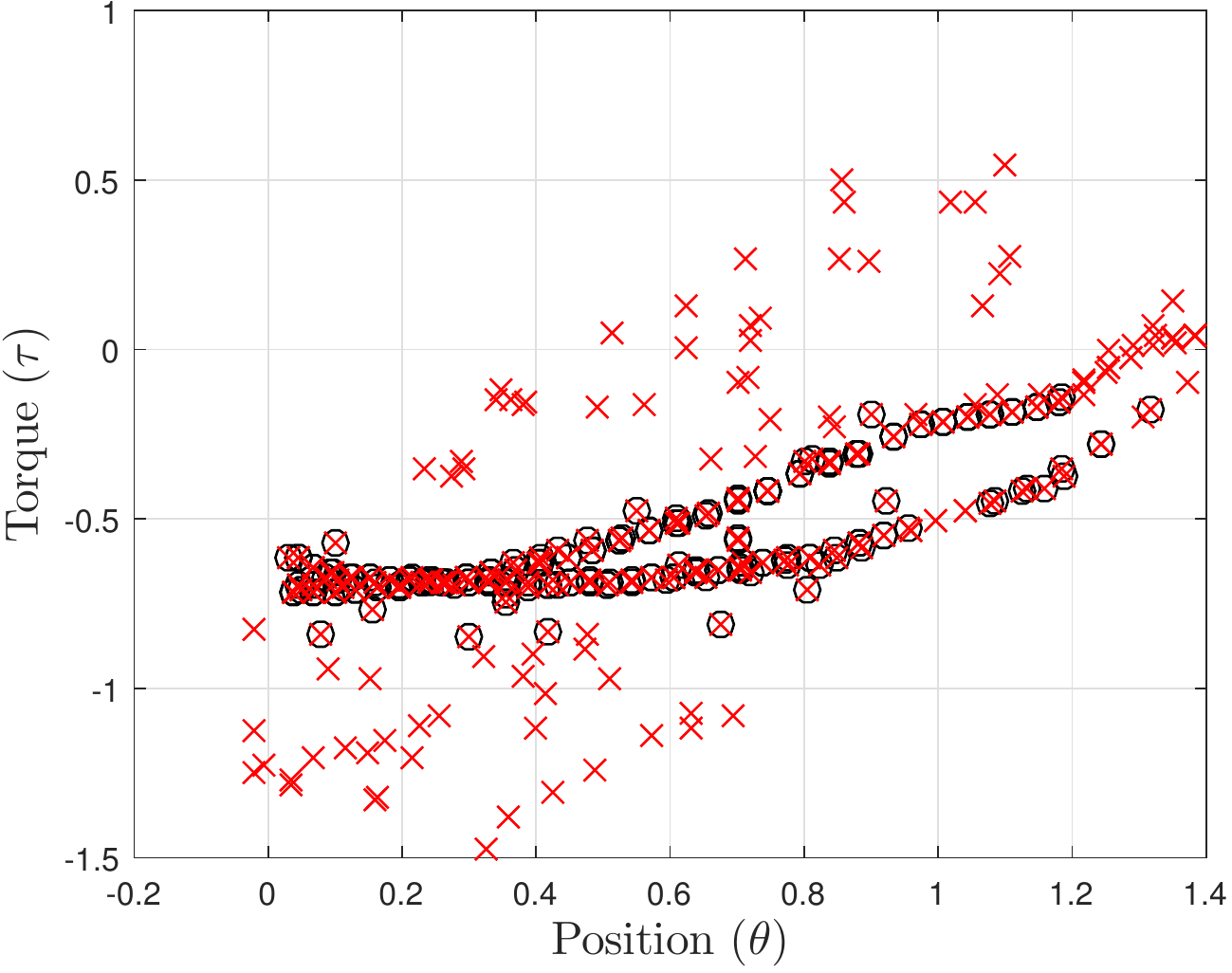}
		\par\end{centering}
	\caption{Rendering of Zero Impedance \label{classification_results}}
\end{figure}

To test the ability to identify with perturbations, the system was manually perturbed during the identification process. The results can be seen in Figure \ref{classification_results}. Using the force sensor data to determine when the system was unperturbed, 192 of these 235 data points were correctly identified as being unperturbed. Of data that had significant external force, 14 of the 92 samples were incorrectly identified as being from the unperturbed system. These results were obtained with the final parameters seen in Table \ref{gpr_param_final}.

\begin{table}[h]
	\centering
	\begin{tabular}{|c|c|}
		\hline 
		Parameter& Value\tabularnewline
		\hline 
		\hline 
		$l$ & $.42$\tabularnewline
		\hline 
		$\sigma_{y}$ & $.05$\tabularnewline
		\hline 
		$\sigma_{n}$ & $.14$\tabularnewline
		\hline 
		$\Sigma_{d}$ & $.07$\tabularnewline
		\hline 
	\end{tabular}
	
	\protect \caption{Final GPR Parameters\label{gpr_param_final}}	
\end{table}

\subsection{Coulomb Friction}
\begin{figure}[H]
	\begin{centering}
		\includegraphics[width=\columnwidth]{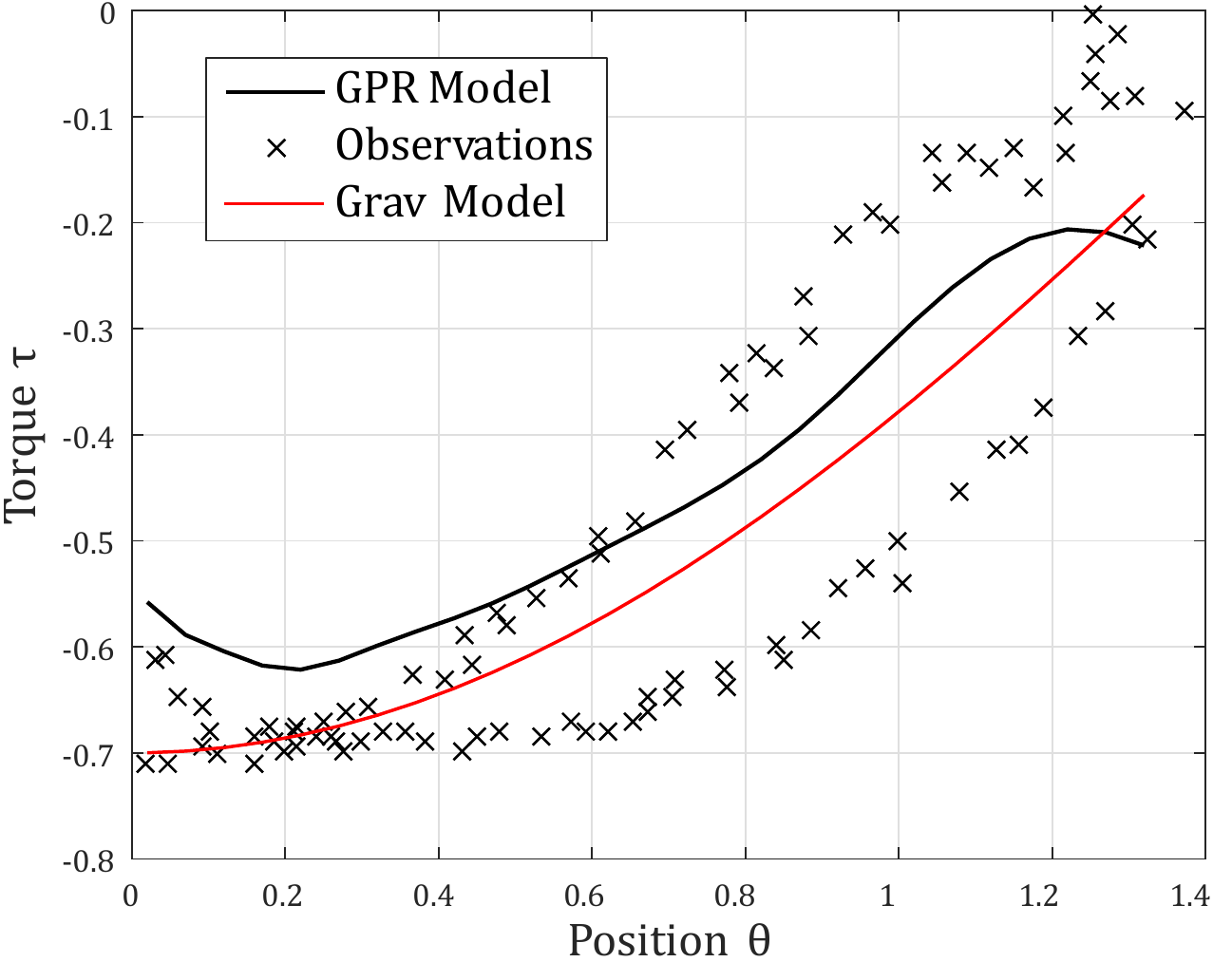}
		\par\end{centering}
	\caption{Regressed Model, Nominal State $x=\left[\theta,\dot{\theta},\ddot{\theta}\right]$\label{model_id_no_sign}}
\end{figure}

\begin{figure}[H]
	\begin{centering}
\		\includegraphics[width=\columnwidth]{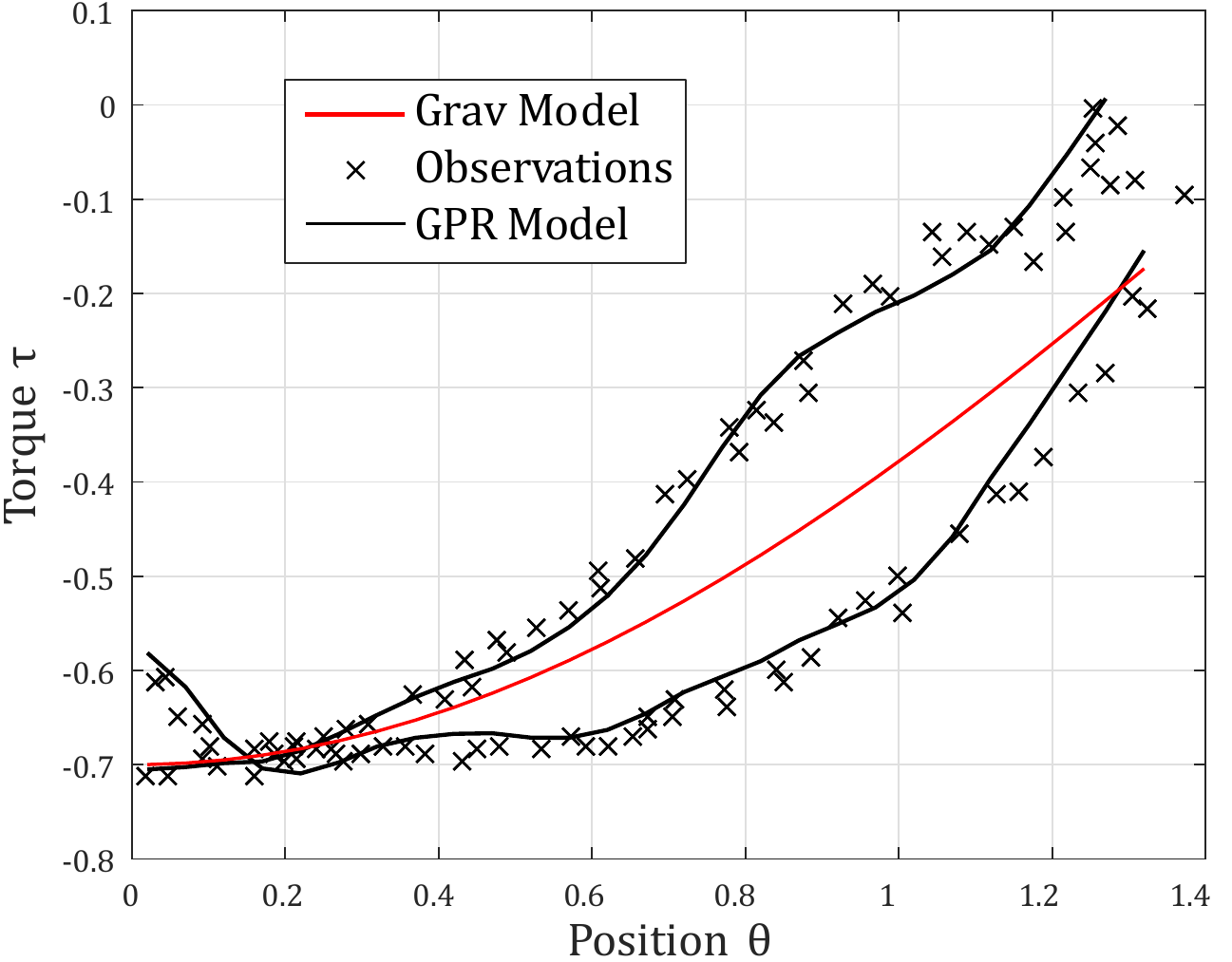}
		\par\end{centering}
	\caption{Regressed Model, Augmented State $x = \left[\theta,\dot{\theta},\ddot{\theta},\mathrm{sgn}(\dot{\theta})\right]$ \label{model_id}}
\end{figure}

To account for the significant coulomb friction in this setup, an additional feature was added to the state representation: $\mathrm{sgn}\left(\dot{\theta}\right)$. Note this violates the conditions for passivity, but as this feature is constant in $\dot{\theta}$, it's ability to inject energy is limited. Seen in Figures \ref{model_id} and \ref{model_id_no_sign}, the discontinuity of coulomb friction can be hard for Gaussian processes to regress (Gaussian processes are continuous and smooth). Introducing this additional feature helps with the separation of these two data sets. 

Also note the load-dependent coulomb friction. This is just one of many position dependent effects which can make analytical models difficult to implement.

\subsection{Impedance Rendering}
\begin{figure}[h]
	\begin{centering}
		\includegraphics[width=\columnwidth]{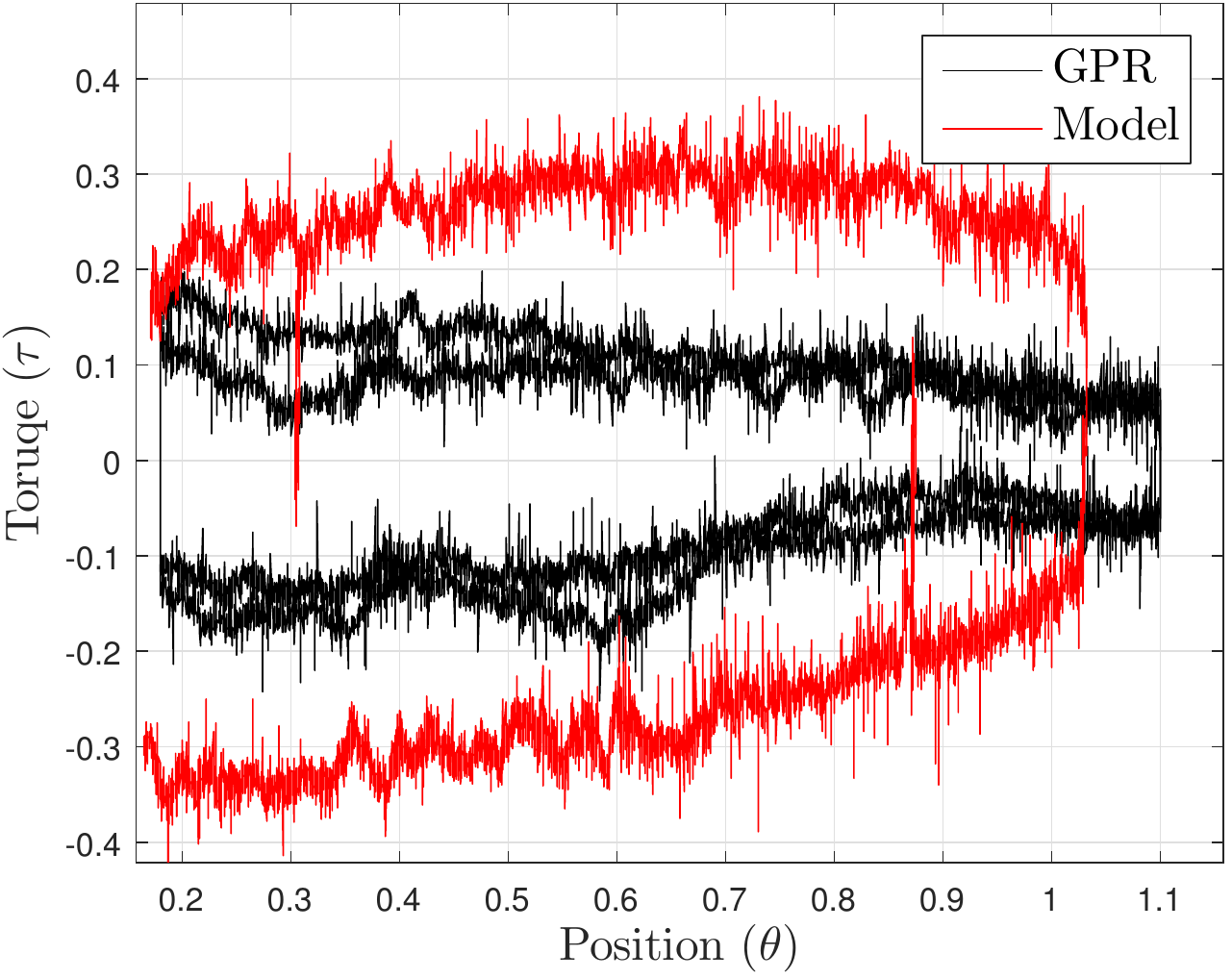}
		\par\end{centering}
	\caption{Rendering of Zero Impedance \label{zero_imp_validation}}
\end{figure}

The identified model is then used for online compensation. In Figure \ref{zero_imp_validation}, the result of rendering a zero impedance can be seen. This system seeks to present no resistance to constant velocity input (the uncompensated inertia means it will resist acceleration). The GPR model, which captures the position-dependent static friction term is better able to compensate for this term. 

\begin{figure}[h]
	\begin{centering}
		\includegraphics[width=\columnwidth]{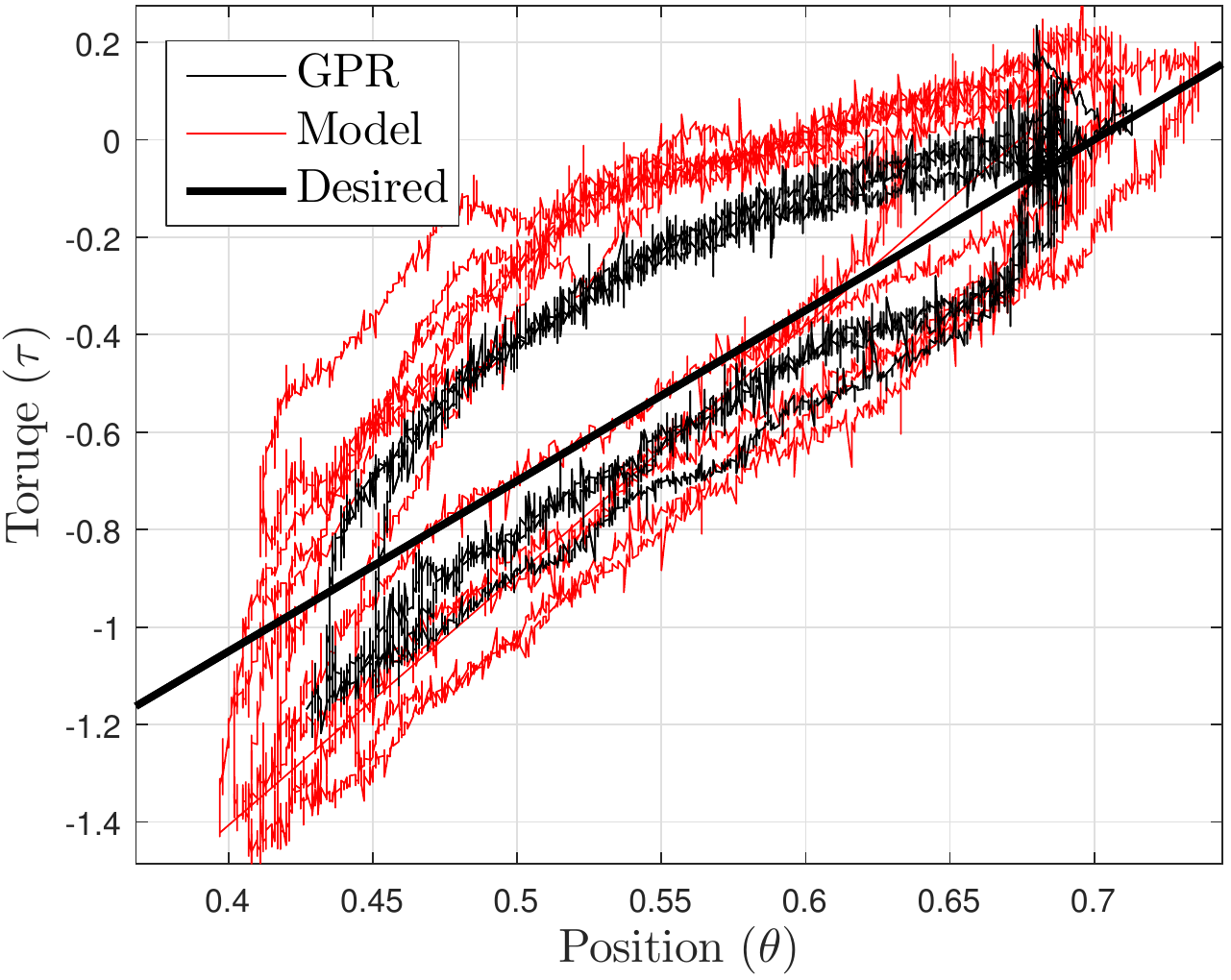}
		\par
	\caption{Rendering of Pure Stiffness $K_{imp}=3.5$\label{imp_kp_35_comparison}}
\end{centering}
\end{figure}

In Figure \ref{imp_kp_35_comparison}, the rendering of a pure stiffness can be seen.  Again, the GPR compensated system provides improved performance. However, the perfect compensation of static friction is not achieved.  

\subsection{Passivity Validation}
\begin{figure}
	\begin{centering}
		\includegraphics[width=\columnwidth]{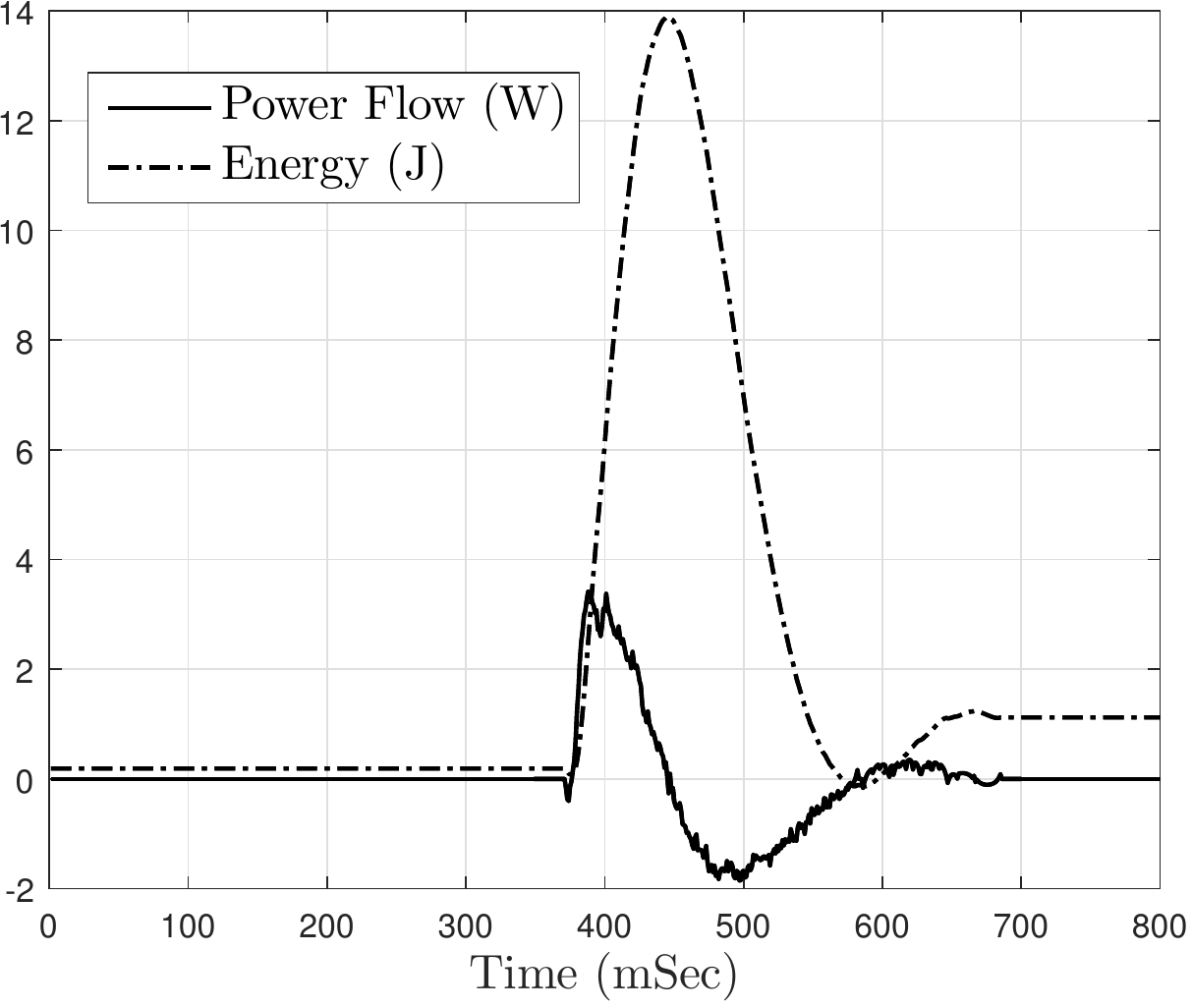}
		\par
	\caption{Power and Energy into Actuator During Environmental Impact\label{power_flow_impact}}
\end{centering}
\end{figure}

Although a rigorous demonstration of passivity experimentally is not feasible (cannot realize all admissable inputs over arbitrary time period), by demonstrating the passivity of the system in a demanding application, passivity can be suggested.  Here, broad-spectrum frequency content is introduced by delivering impulses with a rubber mallet to the load side.  Instantaneous power flowing into the load side from the actuator can be found from $\tau$, which is directly measured and $\dot{\theta}$, which can be found through differentiation and filtering. The instantaneous power flow and total energy into the actuator can be seen in Figure \ref{power_flow_impact}. The actuator absorbs energy through the impact even as it realizes an impedance and feedforward compensation, showing the overall control policy is passive.

\section{Conclusion}

As always, sensing and actuation performance limit the performance which can be achieved by feedforward compensation, however there are other theoretical and practical limit involved for learning inverse dynamic models. When the inverse dynamics lose uniqueness over a state $x$, they cannot be exactly canceled by a function of this state alone.  Common causes of this loss of uniqueness are hidden states, external input and static friction.  To the extent that the system exhibits these behaviors, it cannot be directly compensated by learned inverse dynamics.  

Friction modeling has long been treated separately in robot control, though means of resolving static friction when the system is at rest requires knowledge of the desired direction of motion.  For interactive systems, the desired direction of motion is subject to modification from the environmental input, and thus this feedforward friction compensation can be limited.  Alternative approaches \cite{bernstein2005} can be pursued, but these face other challenges such as noise and gear wear.  Static friction remains a fundamental challenge to modeling and interactivity.

External input can be removed from modeling if it is intermittent as seen here, but more general external input remains a challenge for modeling.  Hidden states are also challenging - the available sensing limits the states which can be inferred, and thus the descriptive power of a model. Whether the current state measurements are sufficient or not (in a practical sense) relies on qualitative assessments undertaken by roboticists, and is one of the fundamental ways in which the designer's a priori knowledge is transfered to autonomous systems.

\bibliographystyle{IEEEtran}
\bibliography{lib}
\end{document}